\documentclass[lettersize,journal,twoside]{IEEEtran}
\usepackage{amsmath,amsfonts}
\usepackage{algorithmic}
\usepackage{array}
\usepackage{textcomp}
\usepackage{stfloats}
\usepackage{url}
\usepackage{booktabs}
\usepackage{verbatim}
\usepackage{graphicx}
\usepackage{arydshln}
\usepackage{adjustbox}
\usepackage{siunitx}
\usepackage{multirow}
\usepackage{capt-of}
\usepackage[caption=false,font=footnotesize,
            labelfont=rm,textfont=rm]{subfig}
\usepackage{booktabs}
\usepackage{adjustbox}
\usepackage{threeparttable}
\usepackage[table]{xcolor}
\usepackage{pifont}
\usepackage{cite}
\usepackage{marvosym}  %

\graphicspath{{./figures/}}

\newcommand{\etal}{\textit{et al}.}

\def\BibTeX{{\rm B\kern-.05em{\sc i\kern-.025em b}\kern-.08em
    T\kern-.1667em\lower.7ex\hbox{E}\kern-.125emX}}
\usepackage{balance}

\usepackage[
    colorlinks=true,
    citecolor=blue,
    linkcolor=blue,
    urlcolor=blue,
    bookmarks=true,
    bookmarksnumbered=true
]{hyperref}
\begin{document}
\title{Real-Time Event-Based LED Tracking and Pose Estimation with PWM-Encoded Markers}
\author{Junkai Niu, Sheng Zhong, Delei Kong, Xin Wang, Kaizhen Sun, Yusen Zhang, Mingzhen Gan, Yi Zhou$^\text{\Letter}$%
\thanks{Junkai Niu, Sheng Zhong, Delei Kong, Xin Wang, Yusen Zhang, Mingzhen Gan and Yi Zhou are with the Neuromorphic Automation and Intelligence Lab (NAIL), School of Artificial Intelligence and Robotics, Hunan University, Changsha, China.}
\thanks{Kaizhen Sun is with the State Key Laboratory for Novel Software Technology, Nanjing University, Nanjing, China.}
\thanks{$\text{\Letter}$~Corresponding author (eeyzhou@hnu.edu.cn).}
}
\markboth{IEEE/ASME Transactions on Mechatronics,~Vol.~XX, No.~XX, October 2026}%
{Niu \MakeLowercase{\textit{et al.}}: Real-Time Event-Based LED Tracking and Pose Estimation with PWM-Encoded Markers}
\setcounter{figure}{-2}
\makeatletter
\g@addto@macro\@maketitle{
\vspace{0ex}
\def\figWidth{0.186\linewidth} 
\centering
\vspace{0.5em}
\setlength\tabcolsep{0pt}
{\small
\centering
{\includegraphics[width=\linewidth]{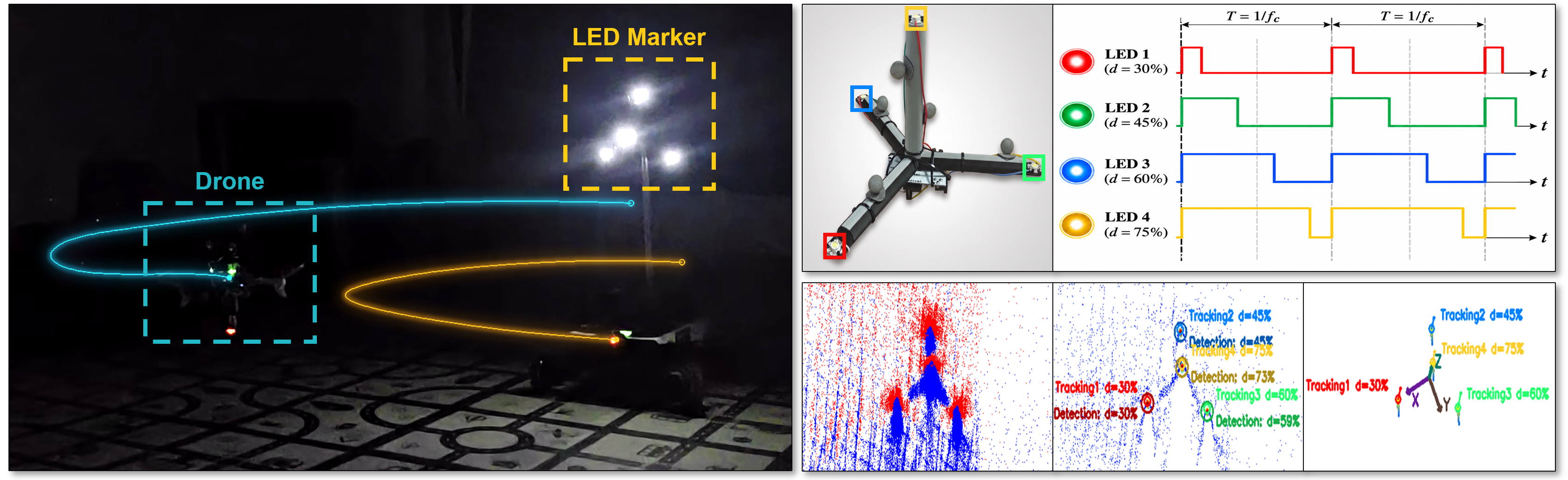}}
}
\vspace{-1em}
\captionof{figure}{We propose a real-time event-based framework for LED tracking and relative pose estimation using pulse-width modulation (PWM)-encoded LED markers.
Left: Drone target-following experiment under varying illumination and HDR conditions. 
Upper right: LED marker and the proposed PWM encoding method, in which LEDs share a common blinking frequency but are distinguished by their duty cycles. 
Lower right: Event map, with positive- and negative-polarity events shown in red and blue, respectively, together with the LED detection and tracking results and the estimated relative pose.}
\label{fig:eyecatcher}

\vspace{-4ex}
}
\makeatother
\maketitle

\begin{abstract}

Event cameras are neuromorphic vision sensors with pixels that independently and asynchronously respond to brightness changes at microsecond resolution, supporting state estimation under rapid motion, motion blur, and high dynamic range (HDR) illumination.
This also enables capture of high-frequency blinking patterns from light-emitting diodes (LEDs).
Based on this, we propose a real-time event-based framework for LED tracking and six-degree-of-freedom (6-DoF) relative pose estimation using pulse-width modulation (PWM)-encoded LED markers.
Multiple LEDs share a common blinking frequency and encode identities through distinct PWM duty cycles, reducing frequency-resource requirements and computational cost.
Our dual-rate asynchronous tracking framework combines low-rate LED detection with high-rate raw-event updates, maintaining robust LED identity association and accurate pixel-level tracking under rapid motion.
LED identities and positions establish reliable 2D-3D correspondences for real-time metric 6-DoF relative pose estimation.
Extensive evaluations on a self-collected dataset demonstrate robustness under large motion flow and transient image-plane LED overlap.
Furthermore, we develop a coordinated robotic platform comprising a quadrotor and a Mecanum-wheeled mobile robot, demonstrating the proposed framework's potential for closed-loop drone guidance through hovering and target-following experiments under time-varying HDR illumination, with onboard deployment on an NVIDIA Jetson Orin NX.
The dataset and implementation will be released to facilitate further research.
\end{abstract}


\section*{Multimedia Material}
\noindent Video: {\small \url{https://youtu.be/n9jH9hwif40}}\\
\vspace{-2.0em}
\section{Introduction}

\IEEEPARstart{R}{obust} and low-latency estimation of the 6-DoF relative pose between rigid bodies is important for robotic manipulation, multi-robot coordination, autonomous landing, and augmented reality~\cite{Gallego2022Survey, Jiao2023LCECalib, Li2026IEICalib, Liu2026AEVINS}.
Conventional solutions often rely on radio-frequency links, motion-capture infrastructure, or frame-based fiducial markers such as AprilTag~\cite{Wang2016AprilTag2} and ArUco~\cite{Garrido2014Aruco}, whose performance can degrade when communication or external infrastructure is unavailable, or when rapid motion and challenging illumination cause motion blur and intermittent observations.
Active LED markers provide a compact and controllable alternative~\cite{Muller2011LED,Censi2013LED}. 
Given a known geometric configuration of LEDs, their position measurements can be used to recover the metric relative pose of the marker~\cite{Ebmer2024ALM}.
Marker identity can be encoded either spatially, through the geometric arrangement of multiple LEDs, or temporally, through their blinking patterns. Compared with spatial coding, temporal coding allows individual LEDs to carry identification information independently, thereby reducing the dependence on complex spatial layouts.
Reliable decoding of such temporal patterns requires accurate sensing of rapid ON-OFF transitions. 
Event cameras are particularly well suited to this task, as they asynchronously capture pixel-wise intensity changes with microsecond-scale temporal resolution, high dynamic range, and low latency~\cite{Lichtsteiner2008DVS,Brandli2014DAVIS,Liu2026AEVINS}, enabling robust perception under high-speed motion and challenging illumination~\cite{Mueggler2014Pose,Wang2024AEB,Niu2025ESVO2,Lyu2026RadarEventFusion}.
For active LEDs, the precise timestamps and polarities of the resulting events directly capture the temporal structure of their blinking signals, supporting event-based LED detection and identification.

Despite these advantages, existing event-based LED identification methods face limitations in achieving robust multi-marker association under asynchronous event streams.
Existing active-marker systems commonly associate LEDs through event counts, inter-event intervals, or distinct blinking frequencies~\cite{Censi2013LED,Ebmer2024ALM,Wang2022Beacons}. 
Assigning a separate frequency to each marker, however, becomes increasingly restrictive as the marker count grows: within a fixed frequency range, increasing the number of markers reduces the spacing between assigned frequencies and can require longer observation intervals for reliable identification~\cite{Bauersfeld2025Mocap}.
Pulse-interval encodes avoid allocating a distinct frequency to each marker by encoding identities through sequences of unequal transition intervals~\cite{Shiba25cvprw}.
This motivates a coding scheme that distinguishes multiple LEDs while retaining a common fixed blinking period for detection.
Moreover, many pipelines accumulate events into fixed-duration windows or pseudo-frames. 
When tracking updates rely on these accumulated detections, the detection interval limits the update rate of position measurements.

These limitations reveal two key challenges: developing a coding scheme that reduces the required frequency resources for reliable multi-marker identification, and designing an asynchronous tracking strategy that combines detection robustness with event-level temporal precision.
We address these challenges with an event-based relative-pose framework based on PWM-encoded LED markers. 
As demonstrated in Fig.~\ref{fig:eyecatcher}, all markers share a common blinking frequency while encoding their identities through duty cycles, thereby reducing the frequency range that needs to be covered by the detector.
Inspired by event-wise blob tracking~\cite{Wang2024AEB}, we adopt a dual-rate asynchronous tracking framework fram that combines periodic detection corrections with high-rate updates from associated raw events.
Finally, the recovered marker identities establish 2D-3D correspondences for accurate perspective-$n$-point (PnP) pose estimation.

The primary contributions of this paper are summarized as follows:
\begin{itemize}

    \item A PWM marker encoding and identification method for event cameras, where multiple LEDs can share the same blinking frequency and be distinguished by their PWM duty cycles, reducing the required frequency resources and consequently lowering the computational cost of marker identification (Sec.~\ref{sec:detection}).

    \item A dual-rate asynchronous tracking framework that combines low-rate LED detections with high-rate updates from associated raw event data for robust marker tracking.
    By integrating both measurements into a unified state estimation framework, the proposed method achieves accurate and responsive tracking under scenarios with large optical flow (Sec.~\ref{sec:tracking}).

    \item A real-time 6-DoF relative pose estimation framework based on the identification and tracking of PWM-encoded LED markers, which can be used for drone guidance in time-variant HDR environments.
Extensive experiments on a self-collected dataset and drone flight experiments demonstrate its effectiveness and practical performance, and the dataset and implementation will be released to facilitate further research (Sec.~\ref{sec:pose}).
\end{itemize}

\emph{Outline}: 
The remainder of the paper is organized as follows. 
Sec.~\ref{sec:related work} reviews active LED coding and identification, as well as event-based marker tracking and pose estimation.
Sec.~\ref{sec:method} introduces the proposed PWM detection, asynchronous tracking, and pose-estimation pipeline. 
Sec.~\ref{sec:evaluation} presents the experimental evaluation, followed by the conclusion in Sec.~\ref{sec:conclusion}.

\section{Related Work}\label{sec:related work}

Event-based optical markers have been studied for tracking, localization, and communication~\cite{Tofighi2026Survey}.
For active-marker pose estimation, LED identification and image-plane tracking are two important components.
The former establishes marker identities for 2D-3D correspondence, while the latter provides temporally consistent visual measurements for geometric pose recovery.

\textbf{Active LED Encoding and Identification.}
Existing active-marker systems commonly distinguish LEDs by their blinking frequencies or by encoded blinking patterns.
Early systems demonstrate LED tracking and low-latency localization using temporally modulated light sources~\cite{Muller2011LED,Censi2013LED}.
Chen \etal ~\cite{Chen2020VLP} extend this direction to visible-light positioning by identifying multiple flickering LEDs, while Salah \etal ~\cite{Salah2022Localization} combine active-landmark measurements with inertial sensing for relative localization.
More recently, Ebmer \etal ~\cite{Ebmer2024ALM} exploit distinct LED frequencies and a known marker geometry to estimate the 6-DoF relative pose.
Bauersfeld and Scaramuzza~\cite{Bauersfeld2025Mocap} adopt a similar identification principle for individually mounted point-source LEDs, whose measured frequencies determine the correspondences used in PnP pose estimation.
In these frequency-encoded systems, periodic blinking provides a detection cue and an identity cue.
When distinct blinking frequencies are assigned to individual LEDs, increasing the number of LEDs within a fixed frequency range reduces the frequency spacing between them, potentially requiring longer observation windows for reliable frequency discrimination.

An alternative manner is to encode information through temporal sequences.
Wang \etal~\cite{Wang2022Beacons} develop smart visual beacons with LED modulation and event-based demodulation for optical communication.
Howard and Hirakawa~\cite{Howard2025Timing} address sensor-induced timing mismatch through dynamic timing correction for LED signal decoding.
For marker identification, Shiba \etal~\cite{Shiba25cvprw} introduce the E-VLC dataset with an interval-based protocol, where sequences of unequal transition intervals encode identities for camera localization.
This avoids assigning a distinct frequency to each LED, but requires resolving the intervals within the transmitted sequence.
Kitade \etal ~\cite{Kitade2024Bicode} propose Bicode, which combines a blinking LED for long-range identification with a spatial 2D marker for short-range pose estimation.
Su \etal ~\cite{Su2025Motion} instead use screen-displayed dynamic visual markers together with event-based tracking to support localization and optical communication under camera motion.
These approaches suggest an alternative design space in which LED identity is encoded independently of the carrier frequency, while a common periodic signal can still be retained for robust detection.

\begin{table}[t]
    \centering
    \caption{Comparison of Event-Based Tracking Methods}
    \label{tab:tracking_comparison}

    \renewcommand{\arraystretch}{1.15}
    \setlength{\tabcolsep}{3pt}

    \resizebox{\linewidth}{!}{%
    \begin{tabular}{lcccc}
        \toprule
        \multirow{2}{*}{\textbf{Method}} &
        \textbf{LED} &
        \textbf{Asynchronous} &
        \textbf{High-Speed} &
        \textbf{Real-Time} \\
        &
        \textbf{Identification} &
        \textbf{Tracking} &
        \textbf{Validation} &
        \textbf{($\geq 0.7$ M events/s)} \\
        \midrule

ACE~\cite{Alzugaray2018ACE}
& \ding{55} & \ding{51} & \ding{55} & \ding{55} \\

HASTE~\cite{Alzugaray2020HASTE}
& \ding{55} & \ding{51} & \ding{55} & \ding{55} \\

EKLT~\cite{Gehrig2020EKLT}
& \ding{55} & \ding{55} & \ding{55} & \ding{55} \\

DeepEvT~\cite{Messikommer2023Tracking}
& \ding{55} & \ding{55} & - & \ding{51} \\

jAER~\cite{jAER2007}
& \ding{55} & \ding{51} & \ding{51} & \ding{51} \\

AEB-Tracker~\cite{Wang2024AEB}
& \ding{55} & \ding{51} & \ding{51} & \ding{55} \\

von Arnim \etal~\cite{vonArnim2024Dynamic}
& \ding{51} & \ding{55} & - & - \\

You \etal~\cite{You2024Temporal}
& \ding{51} & \ding{51} & - & \ding{51} \\

        Ebmer \etal~\cite{Ebmer2024ALM}
        & \ding{51} & \ding{51} & - & - \\

        Bauersfeld \etal~\cite{Bauersfeld2025Mocap}
        & \ding{51} & \ding{55} & - & \ding{51} \\

        \midrule
        \textbf{Ours}
        & \ding{51} & \ding{51} & \ding{51} & \ding{51} \\

        \bottomrule
    \end{tabular}%
    }

    \vspace{0.2em}
    \parbox{\linewidth}{\footnotesize
    \ding{51}: supported;
    \ding{55}: unsupported or below threshold; -: not reported.}
\end{table}

\textbf{Event-Based Marker Tracking.}
The asynchronous output of event cameras has motivated methods that associate and track features directly in the event stream. 
Representative methods in this area are summarized in Tab.~\ref{tab:tracking_comparison}.
Zhu \etal~\cite{Zhu2017Association} formulate event-to-feature matching through probabilistic data association.
Other approaches track corner events~\cite{Alzugaray2018Corner,Alzugaray2018ACE} or maintain patch trajectories through asynchronous hypothesis updates~\cite{Alzugaray2019Hypothesis,Alzugaray2020HASTE}.
Hybrid methods use intensity frames to initialize features and events to track them, including the photometric formulation of EKLT~\cite{Gehrig2020EKLT} and the learned tracker of Messikommer \etal~\cite{Messikommer2023Tracking}.
By updating feature states directly from incoming events, these methods can reduce reliance on frame-rate observations, but generally do not exploit marker-specific identities encoded in the event stream.
Temporal modulation provides an additional cue for active-marker recognition and tracking.
To identify moving beacons, von Arnim \etal~\cite{vonArnim2024Dynamic} combine temporal decoding with tracking driven by sparse optical flow from a spiking neural network.
You \etal~\cite{You2024Temporal} exploit the alternating event polarities generated by blinking LEDs to recognize and track markers through event clustering.
For frequency-encoded LEDs, Ebmer \etal~\cite{Ebmer2024ALM} initialize trackers from accumulated detections and subsequently update their positions with incoming events.
Once initialized, tracking proceeds asynchronously between detection stages, while failed tracks are reinitialized from subsequent detections.
In comparison, Bauersfeld and Scaramuzza~\cite{Bauersfeld2025Mocap} extract LED clusters from short event batches and supply their centroids to a constant-velocity particle filter.
Thus, Ebmer \etal~ enable asynchronous position updates between detection stages, whereas Bauersfeld and Scaramuzza obtain new position measurements at the event-batch rate.
Although these methods demonstrate active-marker tracking in dynamic scenarios, their performance in tracking individual LED under large image-plane motion remains insufficiently characterized.

To track point-like targets under large optical flow, Wang \etal~\cite{Wang2024AEB} propose AEB-Tracker, which combines spatial event association with an extended Kalman filter to update target states from individual raw events.
This enables asynchronous position and motion updates directly from incoming events.
However, AEB-Tracker does not explicitly exploit marker-specific identities encoded in temporal blinking patterns, and its eight-state EKF incurs relatively high per-event computational cost compared with lightweight state models.
These limitations motivate a tracking framework that exploits marker-specific identities while maintaining accurate and efficient image-plane tracking under fast motion.

Unlike frequency-encoded active markers, our framework assigns marker identities through distinct PWM duty cycles while using a common blinking frequency for all LEDs, reducing the need for separate frequency resources. 
Moreover, we integrate periodic detection-based corrections with asynchronous event-level tracking to maintain high-rate image-plane state updates between detection cycles. 
By combining resource-efficient marker identification with high-temporal-resolution tracking, our framework establishes stable 2D-3D correspondences for 6-DoF relative pose estimation under fast motion and temporary LED overlap.

\section{Methodology}\label{sec:method}

\subsection{System Overview}

\begin{figure*}[t]
\centering
\includegraphics[width=1.0\linewidth]{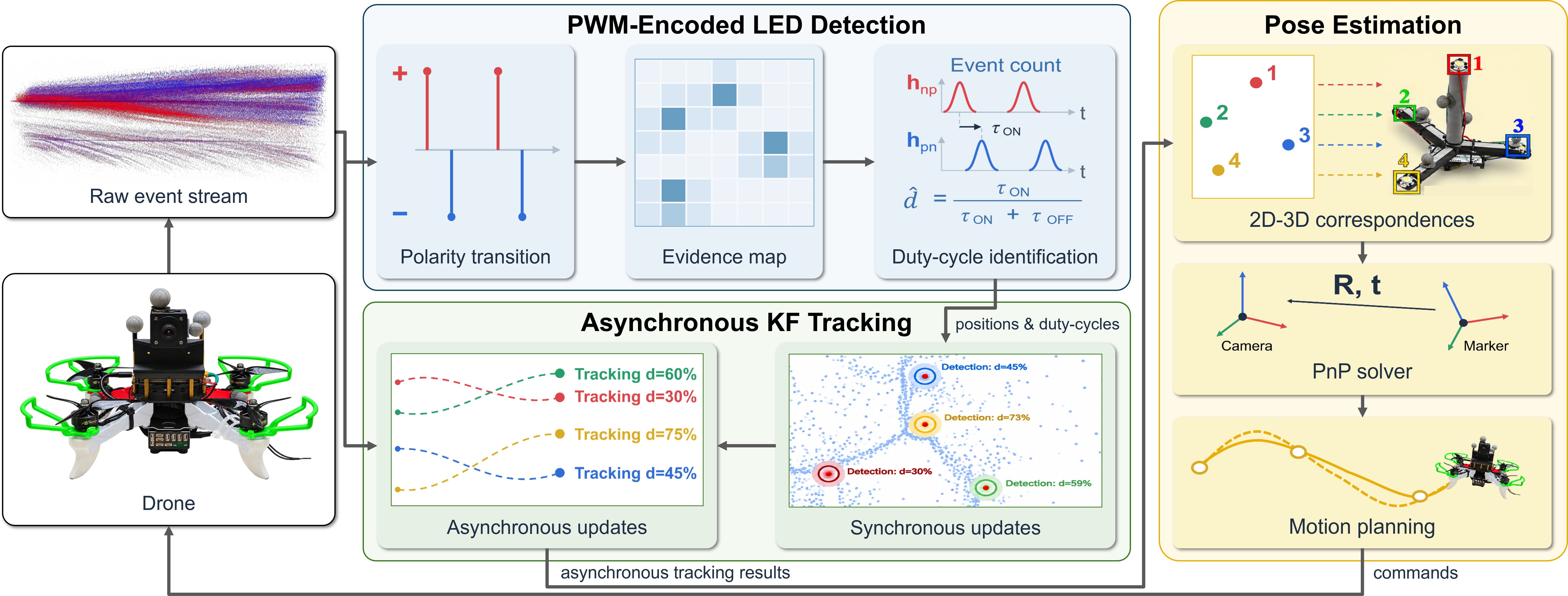}
\caption{\textit{Overview of the proposed event-based relative pose estimation framework.}
The system follows a detect-track-estimate architecture, consisting of PWM-encoded LED detection and identification, asynchronous Kalman filter tracking, and 6-DoF pose estimation.}
\label{fig:pipeline}
\end{figure*}

The goal of our framework is to estimate the 6-DoF relative pose between an event camera and a marker from raw event streams.
The marker is equipped with $N$ LEDs $\mathcal{B}=\{B_1,\dots,B_N\}$, where each LED $B_i$ has a known 3D position $\mathbf{X}_i^{\mathrm{LED}}$ in the marker coordinate frame and is driven by a PWM signal with a unique duty cycle $d_i$.
The relative pose is represented by a rotation matrix $\mathbf{R}\in SO(3)$ and a translation vector $\mathbf{t}\in\mathbb{R}^3$, which transform points from the marker coordinate frame to the camera coordinate frame.
The proposed framework consists of three modules: PWM-encoded LED detection and identification, asynchronous Kalman filter (KF) tracking, and 6-DoF pose estimation, as shown in Fig.~\ref{fig:pipeline}.

The detection and identification module exploits periodic polarity transitions to construct a confidence map, locate candidate LEDs, and recover their identities from the estimated PWM duty cycles (Sec.~\ref{sec:detection}).
Based on these identified observations, the tracking module maintains an independent Kalman filter for each LED, combining periodic detection corrections with asynchronous updates from associated raw events to estimate its pixel coordinates (Sec.~\ref{sec:tracking}).
Finally, the pose estimation module establishes 2D-3D correspondences from the tracked LEDs and estimates the relative pose using PnP, while outlier rejection and degeneracy checks exclude unreliable estimates (Sec.~\ref{sec:pose}).

\subsection{PWM-Encoded LED Detection}\label{sec:detection}

At each detection cycle, the detection module identifies PWM-encoded LEDs from their characteristic blinking patterns in the raw event stream.
All LEDs share a common blinking frequency, while distinct PWM duty cycles encode their identities.
Unlike frequency- or pulse-interval-based coding methods, this design reduces the frequency range that needs to be covered for detection, thereby lowering the computational cost.

\subsubsection{Polarity Transition Analysis}

A PWM-encoded LED periodically switches between ON and OFF states, producing alternating positive and negative events at the corresponding pixel coordinate on the event camera.
To capture this temporal pattern, we maintain the most recent event polarity at each pixel $(x,y)$ and record the timestamps of consecutive positive-to-negative (pn) and negative-to-positive (np) transitions.
Whenever the polarity of an incoming event differs from the most recently recorded polarity at the same pixel, a new transition is registered.
Since two opposite-polarity transitions span only part of a blinking cycle, the instantaneous frequency is estimated from two consecutive transitions of the same type $\delta\in\{\mathrm{pn},\mathrm{np}\}$:
\begin{equation}
f_{k}(x,y)
=
\frac{1}{t_{\delta}^{(k)}-t_{\delta}^{(k-1)}}.
\end{equation}
We represent each polarity transition as $\mathbf{q}_k=[x,y,t,\delta,f_k(x,y)]$, where $t$ denotes the transition timestamp. These transitions are buffered for subsequent confidence map construction.

\subsubsection{Frequency-Weighted Confidence Map}

At each detection cycle, the polarity transitions $\mathbf{q}_k$ within a trailing temporal window of duration $T_w$ are aggregated into a confidence map $\mathbf{C}\in\mathbb{R}^{H\times W}$.
For each pixel $(x,y)$, the confidence value is computed by frequency-weighted voting:
\begin{equation}
\mathbf{C}(x,y)
=
\sum_{k=1}^{M(x,y)}
w\!\left(f_k(x,y)\right),
\end{equation}
where $M(x,y)$ denotes the number of polarity transitions at pixel $(x,y)$ within the temporal window.
The weighting function $w(f)$ adopts a flat-top Gaussian profile:
\begin{equation}
w(f)
=
\begin{cases}
1, & f\in[f_{\min},f_{\max}], \\[2pt]
\exp\!\left(
-\dfrac{(f-f_{\mathrm{bound}})^2}{2\sigma^2}
\right), & \text{otherwise},
\end{cases}
\end{equation}
where $[f_{\min},f_{\max}]$ denotes the frequency passband of the PWM signals, $\sigma$ controls the decay outside the passband, and $f_{\mathrm{bound}}$ denotes the nearest passband boundary.
Transitions within the passband contribute fully to the confidence map, whereas out-of-band transitions are progressively suppressed according to their frequency deviation.
This soft weighting suppresses irrelevant frequency components while retaining tolerance to small frequency estimation errors near the passband boundaries.

\subsubsection{Candidate Region Extraction and Duty Cycle Identification}

The confidence map $\mathbf{C}$ is normalized by its maximum value, and connected components are extracted by contour tracing as candidate LED regions ${\mathcal{R}_j}$.
For each candidate region, an adaptive sampling grid is then constructed according to the width $w$ and height $h$ of its bounding box, with the number of grid divisions determined by
\begin{equation}
K
=
\max\!\left(
1,
\left\lfloor
\frac{\max(w,h)}{r}
\right\rfloor
\right),
\end{equation}
where $r$ is a spatial-resolution parameter.
Within each candidate region, events are ordered by timestamp, and the negative-to-positive and positive-to-negative polarity transition times are recorded as ${t_{\mathrm{np}}}$ and ${t_{\mathrm{pn}}}$, respectively.
The blinking frequency of each candidate is estimated using the discrete Fourier transform (DFT).
Events within a circular region centered at the candidate are accumulated into a temporal sequence with bin width $\Delta t$, multiplied by a Hann window~\cite{harris1978use}, and transformed into the frequency domain.
The dominant frequency $\hat{f}_{\mathrm{DFT}}$ is determined from the maximum spectral magnitude after excluding the direct-current component. 
The positive and negative event streams are processed independently, and the candidate is retained only when their estimated frequencies satisfy the predefined consistency criterion.

The estimated frequency is used for candidate validation, while the PWM duty cycle is used for LED identification.
Given the estimated frequency $\hat{f}_{\mathrm{DFT}}$, the corresponding blinking period is
\begin{equation}
T
=
\frac{1}{\hat{f}_{\mathrm{DFT}}}.
\end{equation}
The rising-edge and falling-edge transitions are then accumulated into temporal histograms $\mathbf{h}_{\mathrm{np}}[k]$ and $\mathbf{h}_{\mathrm{pn}}[k]$ using a bin width $\delta=T/50$.
For a PWM signal, the time interval between the rising and falling transitions corresponds to the LED ON duration $\tau_{\mathrm{ON}}$, which is estimated from the lag at the peak of the cross-correlation between the two transition histograms:
\begin{equation}
\tau_{\mathrm{ON}}
=
\delta
\cdot
\mathop{\arg\max}\limits_{m}
\sum_k
\mathbf{h}_{\mathrm{np}}[k]
\mathbf{h}_{\mathrm{pn}}[k+m].
\end{equation}
Similarly, the OFF duration $\tau_{\mathrm{OFF}}$ is obtained by reversing the order of the two histograms in the cross-correlation.
The duty cycle is then estimated as
\begin{equation}
\hat{d}
=
\frac{\tau_{\mathrm{ON}}}
{\tau_{\mathrm{ON}}+\tau_{\mathrm{OFF}}}.
\end{equation}
The estimate is accepted only when $\tau_{\mathrm{ON}}+\tau_{\mathrm{OFF}}$ is consistent with the DFT-estimated period $T$.
Finally, the estimated duty cycle is matched to the predefined duty-cycle set $\mathcal{D}=\{d_1,\dots,d_N\}$:
\begin{equation}
i^{*}
=
\arg\min_i
\left|
\hat{d}-d_i
\right|.
\end{equation}
The identified LED is then associated with its known 3D position $\mathbf{X}_{i^{*}}^{\mathrm{LED}}$ in the marker coordinate frame.

\subsection{Asynchronous Kalman Filter Tracking}\label{sec:tracking}

Inspired by the event-wise tracking strategy of AEB-Tracker~\cite{Wang2024AEB}, we maintain an asynchronous KF for each identified LED and update it with every associated event.
To achieve real-time processing, the filter state is reduced to the pixel coordinates and velocity of each track.
To mitigate the accuracy degradation caused by this state reduction, the LED size and orientation are provided directly by the detection module.
The detection results are first stabilized by a synchronous KF and then fed to the asynchronous KF as lower-uncertainty observations, forming a dual-rate asynchronous tracking framework.

\subsubsection{State and Observation Model}

The KF adopts a constant-velocity state model.
The state of the track for LED $B_i$ at update step $k$ is defined as
\begin{equation}
\mathbf{x}_{i,k}
=
[c_x,c_y,v_x,v_y]^\top,
\end{equation}
where $(c_x,c_y)$ and $(v_x,v_y)$ denote the pixel coordinate and image-plane velocity, respectively.
The state evolves according to
\begin{equation}
\mathbf{x}_{i,k+1}
=
\mathbf{F}(\Delta t)\mathbf{x}_{i,k}
+
\mathbf{w}_{i,k},
\quad
\mathbf{w}_{i,k}
\sim
\mathcal{N}\!\left(\mathbf{0},\mathbf{Q}(\Delta t)\right),
\end{equation}
where $\Delta t$ denotes the elapsed time since the previous update and
\begin{equation}
\mathbf{F}(\Delta t)
=
\begin{bmatrix}
1 & 0 & \Delta t & 0 \\
0 & 1 & 0 & \Delta t \\
0 & 0 & 1 & 0 \\
0 & 0 & 0 & 1
\end{bmatrix}.
\end{equation}

\noindent The unmodeled image-plane acceleration is represented by continuous white noise with power spectral density $q$, which controls the process noise intensity.
The covariance $\mathbf{Q}(\Delta t)$ of the process noise $\mathbf{w}_{i,k}$ is given by
\begin{equation}
\mathbf{Q}(\Delta t)
=
q
\begin{bmatrix}
\frac{\Delta t^3}{3} & 0 & \frac{\Delta t^2}{2} & 0 \\[2pt]
0 & \frac{\Delta t^3}{3} & 0 & \frac{\Delta t^2}{2} \\[2pt]
\frac{\Delta t^2}{2} & 0 & \Delta t & 0 \\[2pt]
0 & \frac{\Delta t^2}{2} & 0 & \Delta t
\end{bmatrix}.
\end{equation}
This formulation naturally scales the process uncertainty with the irregular intervals between asynchronous updates.

The observation model directly measures the 2D pixel coordinate:
\begin{equation}
\mathbf{z}_{i,k}
=
\mathbf{H}\mathbf{x}_{i,k}
+
\boldsymbol{\eta}_{i,k},
\quad
\boldsymbol{\eta}_{i,k}
\sim
\mathcal{N}\!\left(\mathbf{0},\sigma_m^2\mathbf{I}_2\right),
\end{equation}
where $\mathbf{H}=\begin{bmatrix}\mathbf{I}_2 & \mathbf{0}_{2\times2}\end{bmatrix}$ denotes the observation matrix, $\mathbf{z}_{i,k}$ the measured pixel coordinate, and $\sigma_m$ the standard deviation of the position measurement noise in pixels.

\subsubsection{Dual-Rate Asynchronous Tracking Framework}

For the synchronous update, the pixel coordinate of each detected LED is used as the position measurement.
A new track is initialized upon the first identification of an LED, while subsequent measurements are used to correct the corresponding KF state.

For the asynchronous update, each incoming raw event is first associated with an existing LED track.
The KF state is then propagated from the previous update timestamp to the event timestamp, and the event pixel coordinate is used as an immediate position measurement.
This event-wise update provides high-rate position estimates between consecutive synchronous detection updates.
Event-to-track association is performed using an adaptive association radius centered at the predicted pixel coordinate:
\begin{equation}
r_{\mathrm{assoc}}^{(k+1)}
=
\max\!\left(
0.99\,r_{\mathrm{assoc}}^{(k)},
\max\!\left(1.5\,s,r_{\min}\right)
\right),
\end{equation}
where $s$ denotes the size of the corresponding detected LED region and $r_{\min}$ is the minimum association radius.
The association radius gradually decreases over successive updates while remaining bounded by the detected LED size and the predefined minimum value.
An incoming event is associated with the track only if its distance from the predicted pixel coordinate is smaller than $r_{\mathrm{assoc}}$.
The asynchronous update interval $\Delta t$ is further constrained to $[\Delta t_{\min},\Delta t_{\max}]$ to avoid excessively small or large prediction intervals.
By incorporating both synchronous detection measurements and asynchronous event measurements in the same KF, the tracker combines stable periodic corrections with high-rate event-wise updates.

\subsection{6-DoF Relative Pose Estimation}\label{sec:pose}

For the set of active LED tracks $\mathcal{A}$, each identified LED provides a 2D-3D constraint for pose estimation, with the correspondence associated with LED $B_i$ denoted by
\begin{equation}
\left(\mathbf{z}_i,\mathbf{X}_i^{\mathrm{LED}}\right),
\quad i\in\mathcal{A},
\end{equation}
where $\mathbf{z}_i=(u_i,v_i)$ is the current tracked pixel coordinate and $\mathbf{X}_i^{\mathrm{LED}}$ is the corresponding 3D position of LED.
To suppress tracking outliers before pose estimation, we compute the residual between the current track position and its KF prediction,
\begin{equation}
e_i^{\mathrm{KF}}
=
\left\|
\mathbf{z}_i-\hat{\mathbf{z}}_i^{-}
\right\|_2,
\end{equation}
and retain only the tracks satisfying $e_i^{\mathrm{KF}}\leq\delta_{\mathrm{2D}}$.
The resulting valid index set is denoted by $\mathcal{V}$.

When at least four valid correspondences are available, the relative pose $(\mathbf{R},\mathbf{t})$ from the marker frame to the event camera frame is obtained by minimizing the reprojection error
\begin{equation}
(\mathbf{R},\mathbf{t})
=
 \mathop{\arg\min}\limits_{\mathbf{R},\mathbf{t}}
\sum_{i\in\mathcal{V}}
\left\|
\pi\!\left(
\mathbf{R}\mathbf{X}_i^{\mathrm{LED}}
+
\mathbf{t}
\right)
-
\mathbf{z}_i
\right\|_2^2,
\end{equation}
where $\pi(\cdot)$ denotes the camera projection function.
An initial pose is computed using EPnP~\cite{Lepetit2009EPnP} and subsequently refined by \emph{Levenberg-Marquardt} optimization~\cite{Marquardt1963LeastSquares}.
When more than five valid correspondences are available, RANSAC is further applied to reject mismatches from $\mathcal{V}$.

\section{Experiments}\label{sec:evaluation}

In this section, we provide a comprehensive evaluation of the proposed event-based LED tracking and relative-pose estimation framework.
We first introduce the self-collected indoor dataset and experimental setup used in the evaluation (Sec.~\ref{sec:dataset}).
We then evaluate the full system, including LED tracking and 6-DoF trajectory estimation, through comparisons with representative asynchronous event-based trackers (Sec.~\ref{sec:tracking_accuracy}).
Next, we assess the effectiveness and accuracy of the system in drone hovering and following experiments (Sec.~\ref{sec:flight_evaluation}).
Finally, we report the computational performance of the LED detection and identification, asynchronous tracking, and pose-estimation modules (Sec.~\ref{sec:computational_performance}).

\subsection{Dataset}\label{sec:dataset}

The experiments are conducted using a Prophesee EVK4 event camera with a resolution of $1280\times720$ pixels, equipped with a $3.6$-mm infrared band-pass-filtered lens centered at $960$~nm.
The marker consists of four LEDs with an approximately $140$-mm spatial separation, driven at a common PWM frequency of $1$~kHz with duty cycles of $30\%$, $45\%$, $60\%$, and $75\%$, respectively.
A NOKOV Mars-series optical motion-capture system operating at $120$~Hz provides the absolute poses of the camera and the marker, from which the relative pose is computed as groundtruth.
The groundtruth LED pixel coordinates are obtained by projecting their known 3D positions onto the image plane using the measured relative pose and calibrated camera intrinsics.
Using this setup, we collect eight indoor sequences covering diverse motion and challenging tracking conditions, as summarized in Tab.~\ref{tab:dataset_summary}.

\begin{table}[t]
    \centering
    \caption{Summary of the Self-Collected Indoor Dataset}
    \label{tab:dataset_summary}
    \setlength{\tabcolsep}{4.5pt}
    \renewcommand{\arraystretch}{1.2}
    \resizebox{\linewidth}{!}{\begin{tabular}{lccc}
        \toprule
        \textbf{Sequence} 
        & \textbf{Motion Type} 
        & \textbf{ LEDs} 
        & \textbf{Main Characteristics} \\
        \hline

        \textit{single\_led}
        & General motion
        & 1
        & Large optical flow \\

        \textit{rotation\_slow}
        & Rotation
        & 4
        & Slow rotational motion \\

        \textit{rotation\_normal}
        & Rotation
        & 4
        & Moderate rotational motion \\

        \textit{rotation\_fast}
        & Rotation
        & 4
        & Large optical flow \\

        \textit{translation\_z}
        & Translation
        & 4
        & Translation along the optical axis \\

        \textit{6dof\_normal}
        & 6-DoF
        & 4
        & Moderate 6-DoF motion \\

        \textit{6dof\_overlap}
        & 6-DoF
        & 4
        & Temporary LED overlap \\

        \textit{6dof\_overlap2}
        & 6-DoF
        & 4
        & Temporary LED overlap \\
        \bottomrule
    \end{tabular}}
\end{table}

\subsection{Full System Evaluation}\label{sec:tracking_accuracy}

We compare our proposed LED tracker with HASTE~\cite{Alzugaray2020HASTE}, jAER~\cite{jAER2007}, and AEB-Tracker~\cite{Wang2024AEB}.
For convenience, we use ``\textbf{Ours}'' to denote the full proposed LED tracker, and ``\textbf{Ours$^\dagger$}'' to denote the variant that uses only the asynchronous LED tracking module without periodic detection and identification updates.
\textbf{HASTE} is an asynchronous patch-based tracker that evaluates a discrete set of tracking hypotheses against an event-based feature template and updates the feature state on an event-by-event basis.
\textbf{jAER} denotes the Rectangular Cluster Tracker implemented in the jAER software, which asynchronously tracks event clusters in the image plane.
\textbf{AEB-Tracker} models each target as an event blob and combines nearest-neighbor data association with an EKF for asynchronous state estimation.
To isolate LED tracking performance from detection and identification, all methods are initialized with the same groundtruth LED pixel coordinates and corresponding identities at the beginning of each sequence.

\begin{figure*}[t]
    \centering
    \vspace{-1em}
    \includegraphics[width=0.6\textwidth]{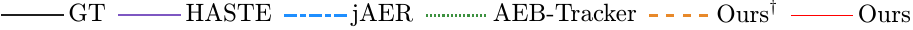}\\[2mm]
    
    \subfloat[\textit{rotation\_slow}\label{fig:tracking_rotation_slow}]
    {\includegraphics[width=0.47\textwidth, height=0.2\textheight]{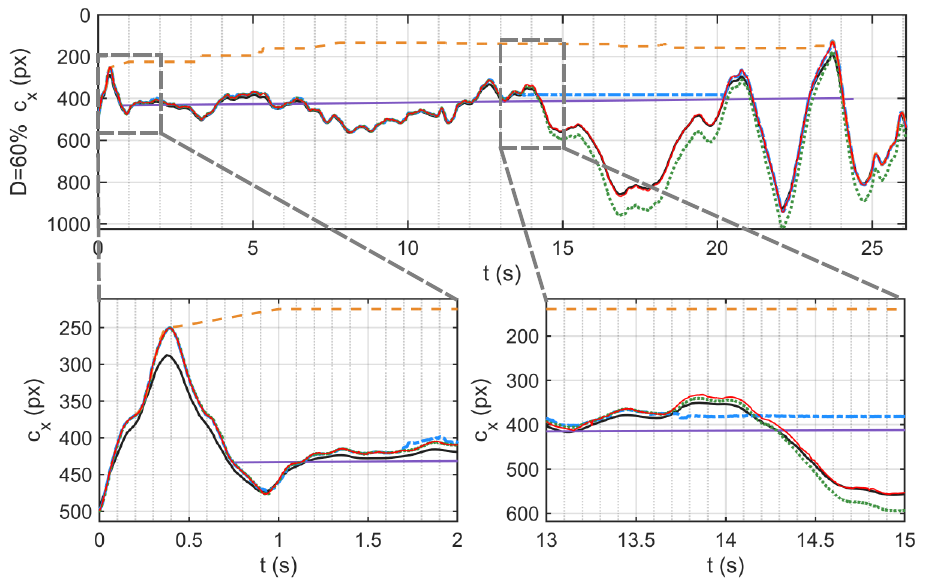}}
    \hspace{2mm}
    \subfloat[\textit{6dof\_normal}\label{fig:tracking_6dof_normal}]
    {\includegraphics[width=0.47\textwidth, height=0.2\textheight]{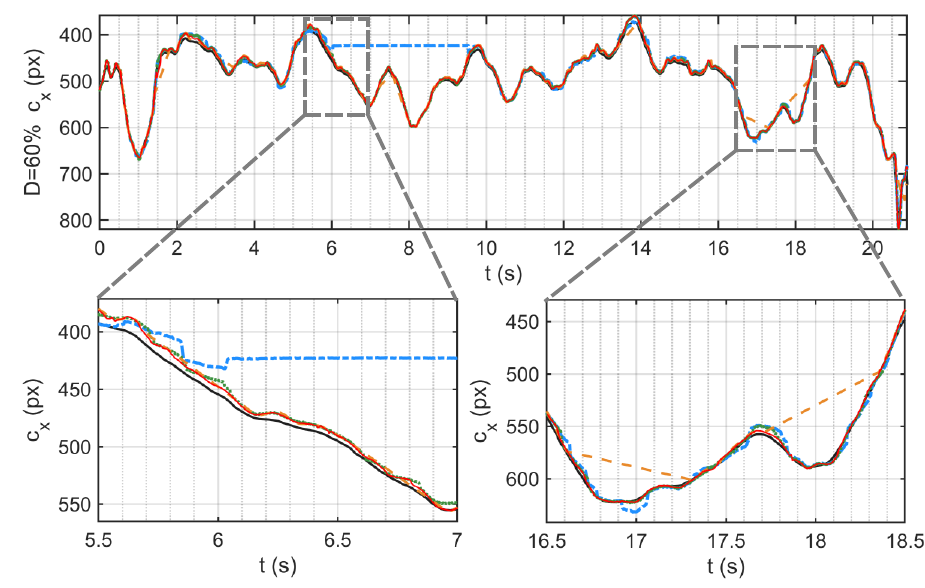}}\\[0mm]
    
    \subfloat[\textit{translation\_z}\label{fig:tracking_translation_z}]
    {\includegraphics[width=0.47\textwidth, height=0.2\textheight]{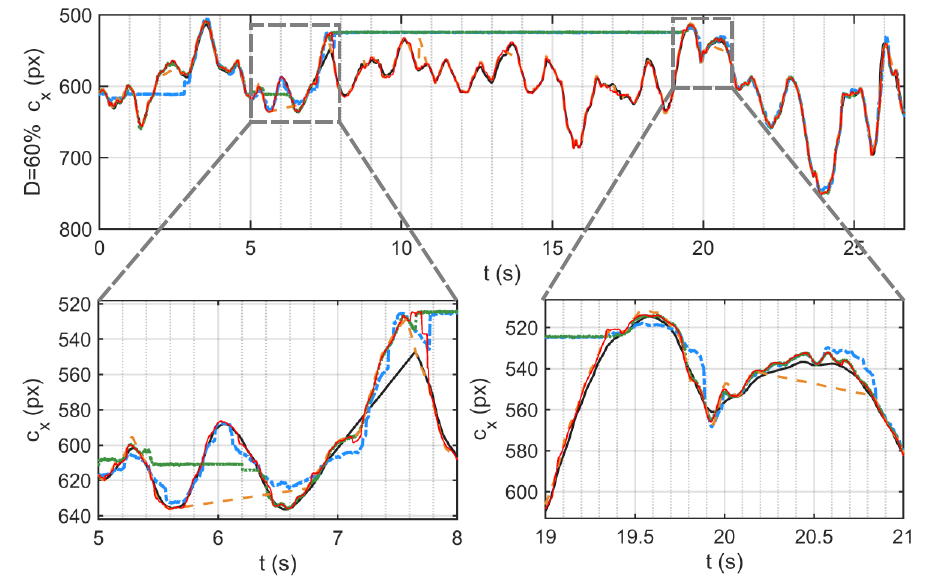}}
    \hspace{2mm}
    \subfloat[\textit{6dof\_overlap}\label{fig:tracking_6dof_overlap}]
    {\includegraphics[width=0.47\textwidth, height=0.2\textheight]{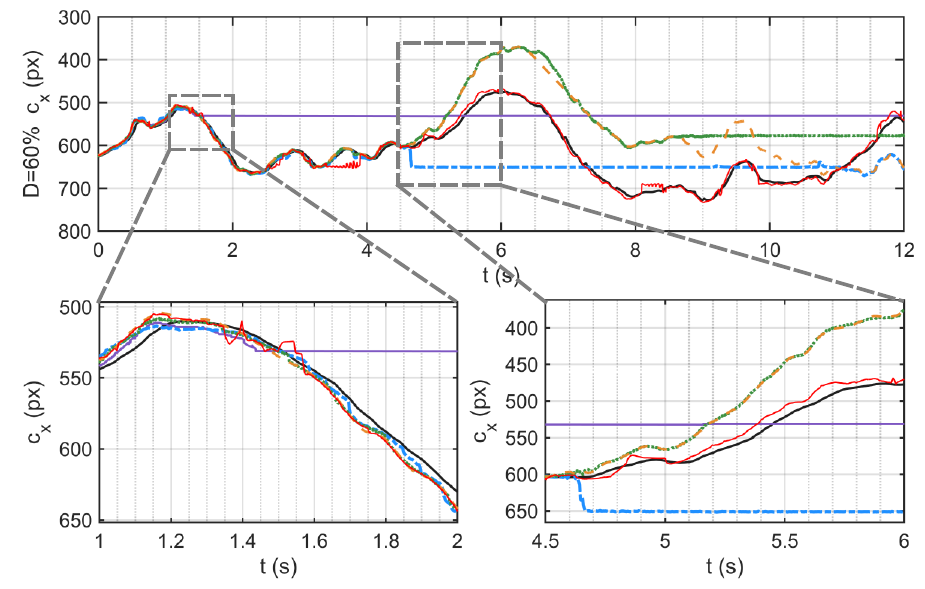}}
    
    \vspace{-0.25em}
    \caption{\textit{LED tracking results for the evaluated methods on part sequences.}
    Each subplot shows the $x$-coordinate tracking results of the $60\%$ duty-cycle LED.}
    \label{fig:tracking_examples}
\end{figure*}

Representative LED tracking results under different motion conditions are shown in Fig.~\ref{fig:tracking_examples}.
In Fig.~\ref{fig:tracking_examples}(a), HASTE fails to establish reliable tracking in most sequences, since the periodic LED blinking continuously changes the event pattern used for template matching.
Figs.~\ref{fig:tracking_examples}(b) and (c) further illustrate that jAER, AEB-Tracker, and Ours$^\dagger$ can be disturbed by event noise, leading to tracking drift or even track loss.
In contrast, the periodic detection and identification updates of our full method provide position corrections that pull the deviated track back toward the corresponding LED observation.
A more challenging case is presented in Fig.~\ref{fig:tracking_examples}(d), where temporary overlap between the LED pixel coordinates can confuse the target association of trackers without detection and identification, whereas our full method can re-establish the correct LED correspondence after the overlap.
These qualitative observations are quantified using the root mean square error (RMSE) between the estimated and groundtruth LED pixel coordinates, as reported in Tab.~\ref{tab:tracking_accuracy}.

\begin{table}[!t]
\begin{minipage}{\linewidth}
\centering
\caption{\label{tab:tracking_accuracy}
Comparison of LED Tracking Results on the Self-Collected Dataset.
[$\mathbf{RMSE}$: px]
Best and second-best results are \textbf{highlighted} and \underline{underlined}, respectively
}
\renewcommand{\arraystretch}{1.11}
\setlength{\tabcolsep}{4pt}
\newcommand{\tabincell}[2]{\begin{tabular}{@{}#1@{}}#2\end{tabular}}
\resizebox{\linewidth}{!}{
\begin{tabular}{lrrrrr}
\toprule
\textbf{Sequence}
& \tabincell{c}{\textbf{HASTE}~\cite{Alzugaray2020HASTE}}
& \tabincell{c}{\textbf{jAER}~\cite{jAER2007}}
& \tabincell{c}{\textbf{AEB-Tracker}~\cite{Wang2024AEB}}
& \textbf{Ours}$^\dagger$
& \textbf{Ours} \\
\midrule

\emph{single\_led}
& failed     & \underline{14.39} & 15.59  & 148.81 & \textbf{13.37} \\

\emph{rotation\_slow}
& 155.15 & 134.73 & 86.78  & \textbf{6.58} & \underline{8.28} \\

\emph{rotation\_normal}
& 156.69 & \underline{8.64} & 59.08  & 32.72  & \textbf{7.93} \\

\emph{rotation\_fast}
& 110.75 & 13.99  & \underline{13.21} & 149.59 & \textbf{13.06} \\

\emph{translation\_z}
& failed     & 65.17  & 32.54  & \underline{10.30} & \textbf{7.12} \\

\emph{6dof\_normal}
& 120.01 & 17.16  & \underline{6.93} & 16.29  & \textbf{6.66} \\

\emph{6dof\_overlap}
& 138.05 & 96.92  & \underline{87.22} & 123.16 & \textbf{28.05} \\

\emph{6dof\_overlap2}
& 284.51 & \underline{95.19} & 118.88 & 161.26 & \textbf{12.57} \\

\midrule
\textbf{Average}
& 160.86 & 55.77 & \underline{52.53} & 81.09 & \textbf{12.13} \\

\bottomrule
\end{tabular}
}
\end{minipage}
\vspace{0.3em}

{\emph{* Averages are computed over the sequences with valid results.}}
\end{table}

As shown in Tab.~\ref{tab:tracking_accuracy}, the proposed method achieves the best LED tracking accuracy on all sequences, with the lowest average RMSE of $12.13$ pixels.
A comparison between AEB-Tracker and Ours$^\dagger$ suggests that the lightweight four-state KF used in our asynchronous tracker trades some standalone tracking accuracy for reduced computational complexity, as reflected by the higher average RMSE of Ours$^\dagger$.
However, when periodic detection and identification are incorporated, the full method substantially outperforms both Ours$^\dagger$ and AEB-Tracker.
This indicates that the proposed detection module provides reliable position and identity corrections that effectively compensate for the reduced standalone accuracy of the lightweight asynchronous tracker and suppress accumulated tracking errors.
The benefit becomes particularly evident under large optical flow and temporary image-plane overlap between LEDs, where the full method maintains consistently low tracking errors while Ours$^\dagger$ degrades markedly.

\begin{table}[!t]
\begin{minipage}{\linewidth}
\centering
\caption{Comparison of 6-DoF Trajectory Results on the Self-Collected Dataset.
[$\mathbf{ATE}$: cm]
Best and second-best results are \textbf{highlighted} and \underline{underlined}, respectively}
\label{tab:pose_ate}
\begin{adjustbox}{max width=\linewidth}
\renewcommand{\arraystretch}{1.11}
\setlength{\tabcolsep}{6pt}
\begin{tabular}{lrrrr}
\toprule
\textbf{Sequence}
& \textbf{jAER}~\cite{jAER2007}
& \textbf{AEB-Tracker}~\cite{Wang2024AEB}
& \textbf{Ours$^\dagger$}
& \textbf{Ours} \\
\midrule
\emph{rotation\_slow}
& 124.23 & 150.62 & \textbf{25.44} & \underline{37.09} \\
\emph{rotation\_normal}
& \underline{13.23} & 159.28 & 78.20 & \textbf{2.97} \\
\emph{rotation\_fast}
& 16.66 & \underline{9.05} & 98.71 & \textbf{4.27} \\
\emph{translation\_z}
& 122.55 & 143.32 & \underline{14.07} & \textbf{13.93} \\
\emph{6dof\_normal}
& 44.77 & \underline{15.23} & 30.86 & \textbf{7.73} \\
\emph{6dof\_overlap}
& \underline{72.37} & 120.59 & 95.28 & \textbf{27.55} \\
\emph{6dof\_overlap2}
& 107.60 & 119.40 & \underline{62.89} & \textbf{15.07} \\
\midrule
\textbf{Average}
& 71.63 & 102.50 & \underline{57.92} & \textbf{15.52} \\
\bottomrule
\end{tabular}
\end{adjustbox}
\end{minipage}
\end{table}

We further evaluate how the differences in LED tracking accuracy affect the downstream 6-DoF relative pose estimation.
As summarized in Tab.~\ref{tab:pose_ate}, our full system achieves the lowest ATE on six of the seven sequences, with an average error of $15.52$~cm.
Compared with Ours$^\dagger$, the substantial reduction in average ATE demonstrates that the periodic detection and identification updates are also critical to the downstream pose estimation.
By correcting LED positions and identities during tracking, our full system maintains more reliable 2D-3D correspondences and prevents accumulated tracking and association errors from propagating into the pose estimates, particularly under fast motion and temporary LED image-plane overlap.

\subsection{Drone Hovering and Target-Following Experiments}\label{sec:flight_evaluation}
\begin{figure*}[t]
    \centering
    \vspace{-1em}

    \includegraphics[width=0.8\textwidth]{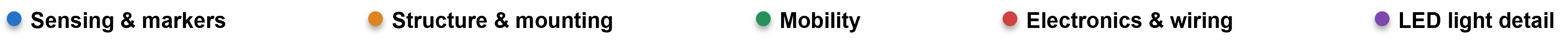}\\
    \vspace{-0.8em}
    \subfloat[\label{fig:mecanum_platform}]
    {
        \includegraphics[height=0.342\textwidth]
        {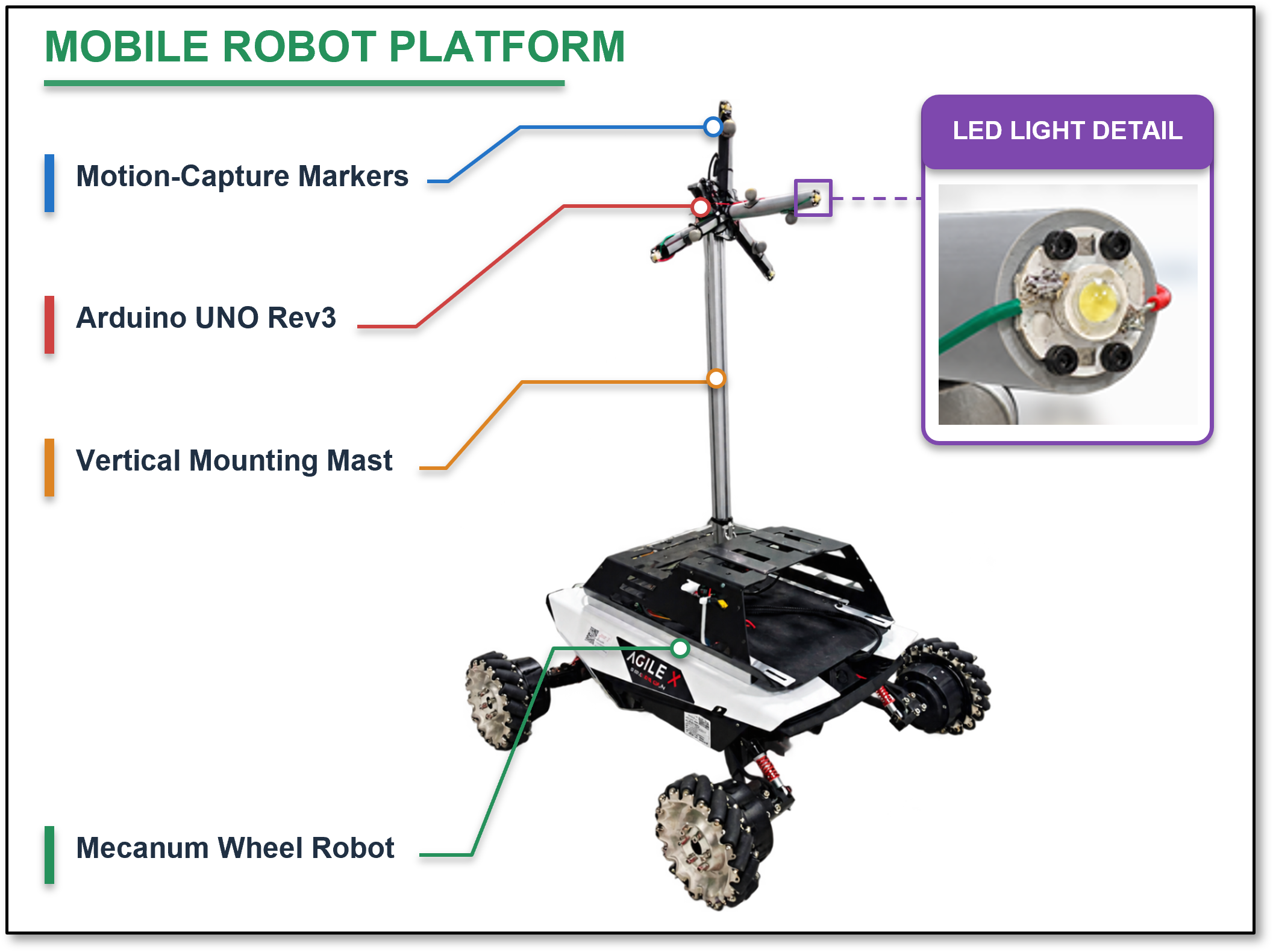}
    }
    \hspace{-1.0em}
    \subfloat[\label{fig:quadrotor_platform}]
    {
        \includegraphics[height=0.342\textwidth]
        {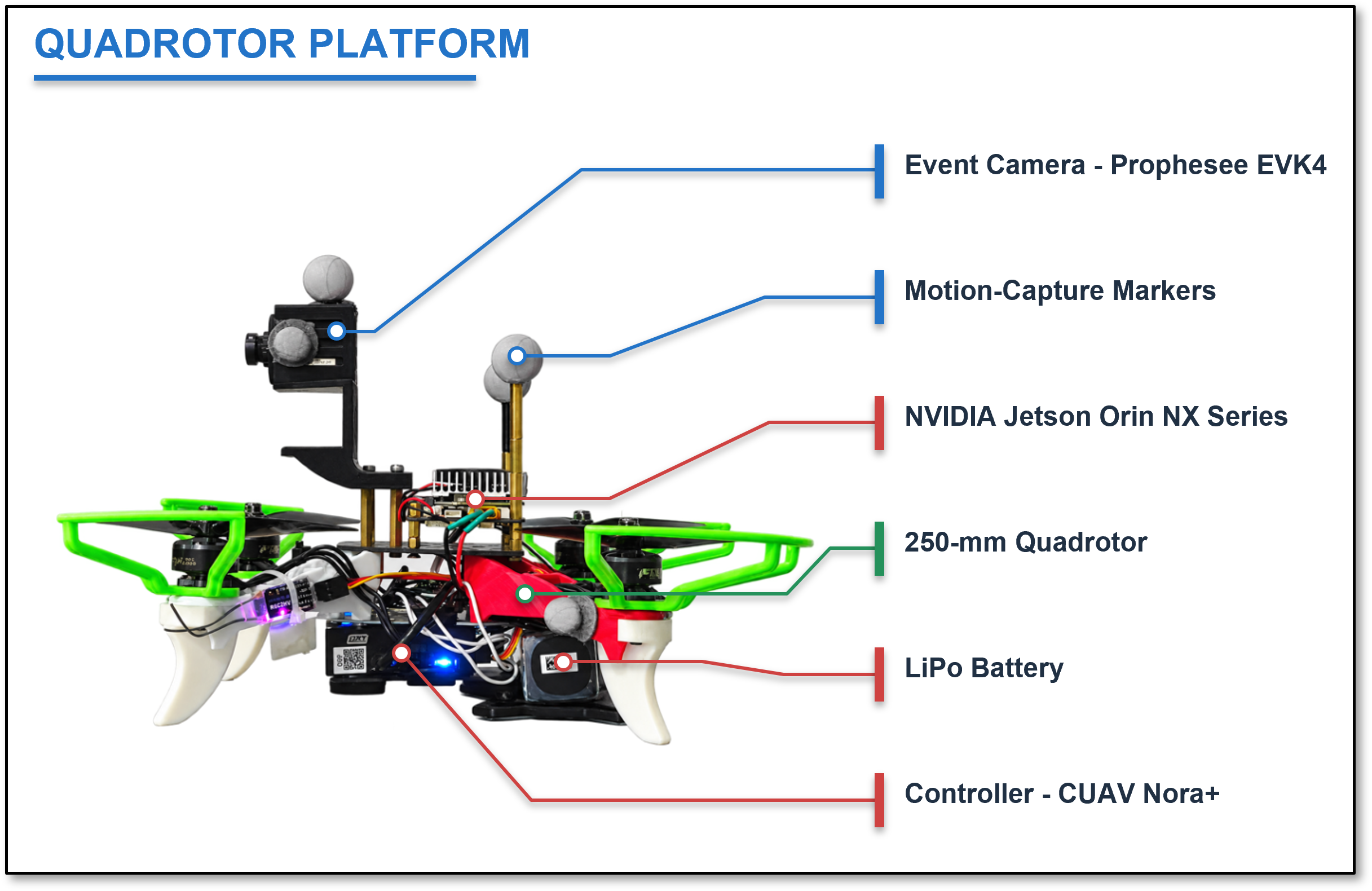}
    }

    \vspace{-0.25em}
    \caption{\textit{Experimental platforms for relative-pose-based drone guidance.}
    (a) The mecanum-wheeled car carrying the PWM-encoded LED marker.
    (b) The quadrotor equipped with a Prophesee EVK4 event camera
    and an NVIDIA Jetson Orin NX.}
    \label{fig:experimental_platforms}
\end{figure*}

\begin{figure*}[t]
\centering
\vspace{-0em}
\includegraphics[width=0.42\textwidth]{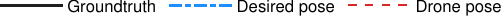}\\[-1mm]

\subfloat[Stationary\label{fig:drone_hovering}]{
    \includegraphics[width=0.315\textwidth]{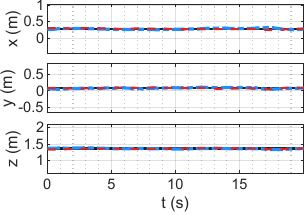}
}
\hfill
\subfloat[Linear motion\label{fig:drone_following_line}]{
    \includegraphics[width=0.308\textwidth]{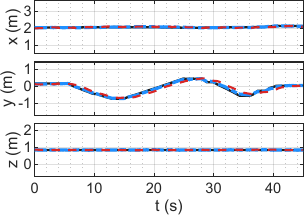}
}
\hfill
\subfloat[Circular motion\label{fig:drone_following_circle}]{
    \includegraphics[width=0.308\textwidth]{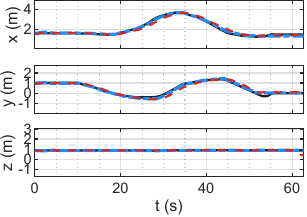}
}
\vspace{-0.25em}
\caption{Drone relative-position control using the estimated relative pose under three marker motion conditions: stationary, linear, and circular motion.}
\label{fig:drone_flight}
\end{figure*}

To assess the practical performance of the proposed system, we conduct drone hovering and target-following experiments using a quadrotor equipped with a Prophesee EVK4 event camera and an NVIDIA Jetson Orin NX.
The estimated relative pose is used to generate the desired drone position for position-feedback control with respect to a PWM-encoded LED marker mounted on a Mecanum-wheeled car, as shown in Fig.~\ref{fig:experimental_platforms}.
In the hovering experiment shown in Fig.~\ref{fig:drone_hovering}, the desired and actual drone trajectories closely follow the groundtruth trajectory, with ATEs of $2.59$~cm and $2.60$~cm, respectively.
In the two target-following experiments, the actual drone trajectories exhibit larger errors than the desired trajectories, mainly due to control latency during target motion.
During the linear-motion experiment in Fig.~\ref{fig:drone_following_line}, the desired trajectory generated from the estimated relative pose achieves an ATE of $4.26$~cm, while the actual drone trajectory exhibits an ATE of $13.29$~cm.
For the circular-motion experiment in Fig.~\ref{fig:drone_following_circle}, conducted under varying illumination and HDR conditions, the desired and actual drone trajectories achieve ATEs of $9.66$~cm and $14.91$~cm, respectively.
The increased flight errors during target motion are mainly attributed to control latency.
Overall, these experiments demonstrate that the proposed event-based relative pose estimation system can support closed-loop drone hovering and target-following control, including under challenging illumination conditions.

\subsection{Computational Performance}\label{sec:computational_performance}

We evaluate the computational performance of the proposed framework on a desktop computer equipped with an Intel Core i7-14700 CPU, without GPU acceleration, excluding visualization and file I/O from all runtime measurements.
Runtime statistics are collected on the \textit{rotation\_normal} sequence over intervals where all evaluated methods maintain stable tracking.
Since all methods operate asynchronously on the raw event stream, we compare the computational cost of their event-wise tracking updates.
To exclude the influence of events rejected by the respective gating or association mechanisms, only events accepted for tracking are included in the statistics.
Tab.~\ref{tab:tracking_runtime} reports the average computational cost per accepted event.
Our method achieves the lowest cost, requiring approximately $15.3$~ns per event.
Compared with the 8-state EKF used by AEB-tracker, our tracker maintains only 2D position and velocity in a four-state linear KF, substantially reducing the cost of each update.

\begin{table}[t]
\begin{minipage}{0.98\linewidth}
    \centering
    \caption{Computational Cost of the Asynchronous Tracking Update for Different Event-Based Trackers}
    \label{tab:tracking_runtime}
    \renewcommand{\arraystretch}{1.2}
    \setlength{\tabcolsep}{5pt}
    \begin{tabular}{llr}
        \toprule
        Method & Asynchronous update & Time / event [ns] \\
        \midrule
        HASTE~\cite{Alzugaray2020HASTE}       & Patch update           & 217.0  \\
        jAER~\cite{jAER2007}        & Rectangular Cluster Tracker update & 32.1   \\
        AEB-tracker~\cite{Wang2024AEB} &  8-state EKF update              & 2139.0 \\
        Ours        &  4-state KF update               & \textbf{15.3}   \\
        \bottomrule
    \end{tabular}
    \end{minipage}
\end{table}

\begin{table}[!t]
\centering
\begin{minipage}{0.98\linewidth}
\centering
\caption{Computational Performance of the Proposed Framework on the Jetson Orin NX}
\label{tab:system_throughput}
\renewcommand\arraystretch{1.2}
\begin{adjustbox}{max width=\linewidth}
\setlength{\tabcolsep}{6pt}
\begin{tabular}{llrr}
\toprule
\textbf{Component}
& \textbf{Function}
& \textbf{Time}
& \textbf{Rate} \\
\midrule

\multirow{4}{*}{Asynchronous}
& Event Transformation
& 9.44 ns/event
& 105.9 M events/s \\

{~}
& Polarity Transition Analysis
& 35.71 ns/event
& 28.0 M events/s \\

{~}
& LED Tracking
& 11.24 ns/event
& 89.0 M events/s \\

\cdashline{2-4}\vspace{-3mm}\\
{~}
& \textbf{Overall}
& \textbf{56.88 ns/event}
& \textbf{17.6 M events/s} \\

\midrule

\multirow{3}{*}{Synchronous}
& Candidate Region Extraction
& 1.69 ms
& 591.7 Hz \\

{~}
& Duty Cycle Identification
& 4.55 ms
& 220.0 Hz \\

\cdashline{2-4}\vspace{-3mm}\\
{~}
& \textbf{Overall}
& \textbf{6.24 ms}
& \textbf{160.2 Hz} \\

\midrule

Pose Estimation
& 6-DoF Pose Estimation
& $\sim$0.53 ms
& $\sim$1.9 kHz \\

\bottomrule
\end{tabular}
\end{adjustbox}
\end{minipage}
\end{table}

To evaluate the real-time capability of the complete framework on an onboard computing platform, we deploy the system on an NVIDIA Jetson Orin NX and report its throughput in Tab.~\ref{tab:system_throughput}.
The asynchronous event-processing module, dominated by polarity-transition analysis, sustains approximately $17.6$~M events/s, exceeding the average input rate of $13.22$~M events/s on the evaluated sequence.
The synchronous LED detection and identification module, whose primary computational cost comes from duty-cycle identification, operates at approximately $160.2$~Hz, well above the $20$~Hz detection rate used by the system.
The relative pose estimation module further supports approximately $1.9$~kHz.
Together, these results confirm that all major components satisfy their respective real-time processing requirements on the Jetson Orin NX.

\section{Conclusion}
\label{sec:conclusion}

We present an event-based real-time relative pose estimation framework using PWM-encoded LED markers.
The proposed PWM encoding scheme allows multiple LEDs to share a common blinking frequency while encoding their identities through distinct duty cycles, thereby reducing the required frequency resources while lowering the computational cost.
To achieve robust marker tracking, we develop a dual-rate asynchronous tracking framework that combines periodic detection and identification corrections with high-rate updates from raw events.
The tracked LED identities and pixel coordinates are then used to establish reliable 2D-3D correspondences for relative pose estimation.
Extensive evaluations on self-collected dataset demonstrate the robustness and effectiveness of the proposed framework compared with representative asynchronous event-based trackers, particularly under large motion flow and temporary image-plane overlap between LEDs.
Drone hovering and target-following experiments with the proposed framework running on an NVIDIA Jetson Orin NX demonstrate its real-time capability and potential for drone guidance under time-varying HDR illumination.

\bibliographystyle{IEEEtran}
\bibliography{references}

\vspace{-1.0em}

\begin{IEEEbiography}[{\includegraphics[width=1in,height=1.25in,clip,keepaspectratio]{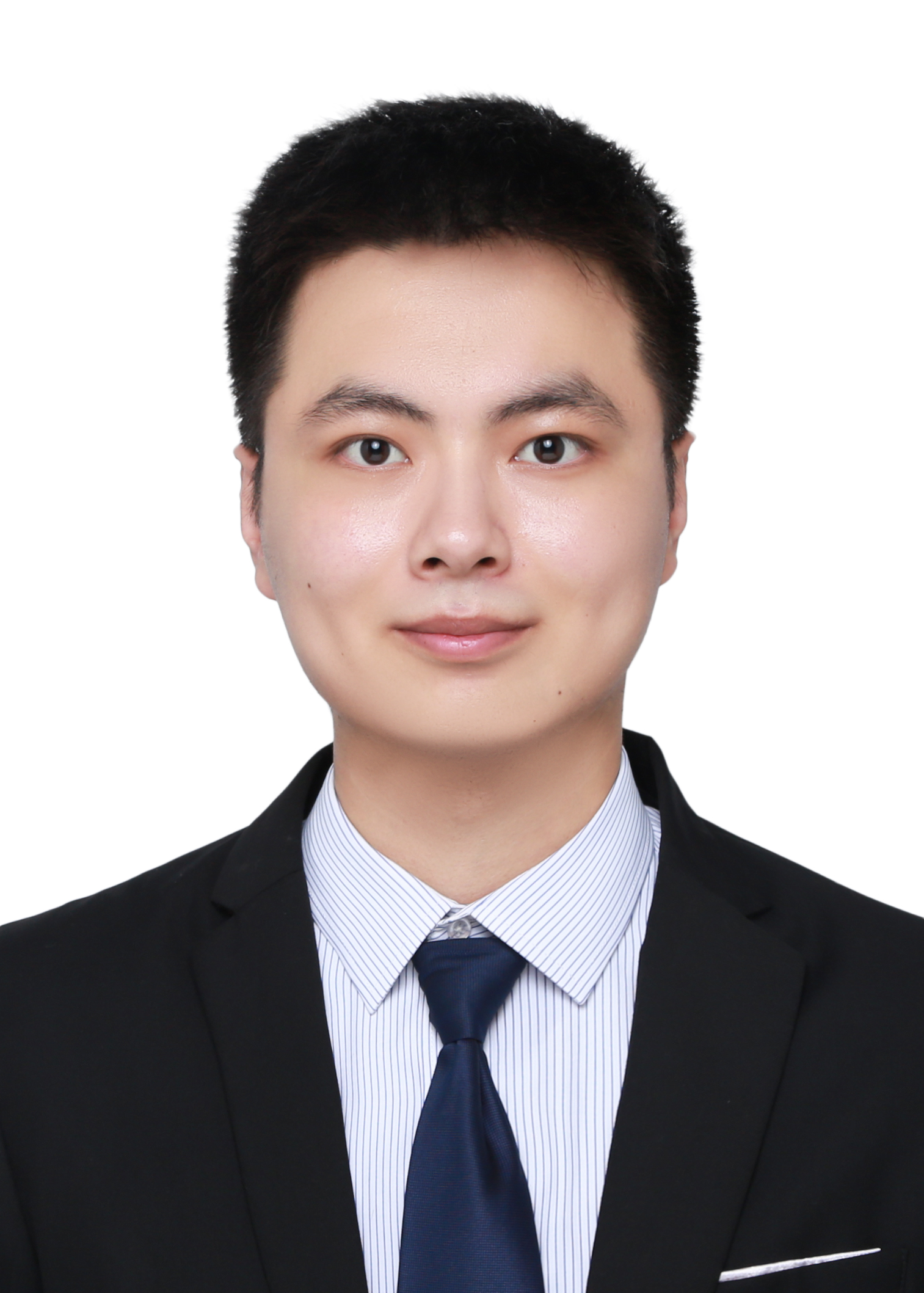}}]{Junkai Niu}
received the Master's degree in Aerospace Science and Technology from the University of Electronic Science and Technology of China, Chengdu, China in 2023.
He is currently working toward the Ph.D. degree in the Neuromorphic Automation and Intelligence Lab (NAIL) with School of Robotics at Hunan University.
His research interests primarily focus on event-based visual odometry/simultaneous localization and mapping.
\end{IEEEbiography}

\begin{IEEEbiography}[{\includegraphics[width=1in,height=1.25in,clip,keepaspectratio]{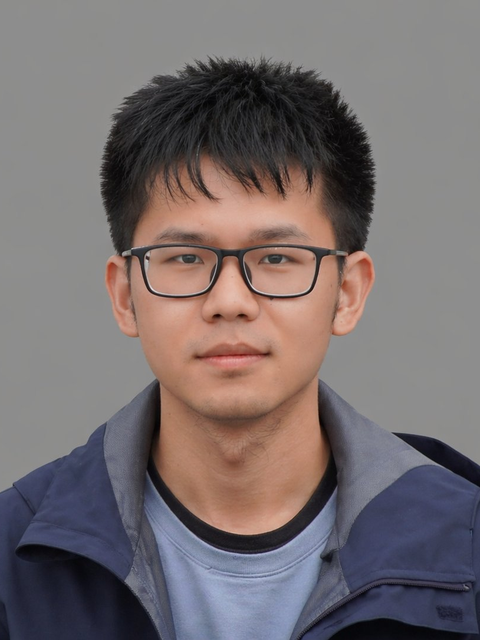}}]{Sheng Zhong}
received the B.Eng. degree in Robotics Engineering from Zhejiang University, Hangzhou, China, in 2023. 
He is currently working toward the Ph.D. degree in the Neuromorphic Automation and Intelligence Lab (NAIL) with School of Robotics at Hunan University.
His research interests primarily focus on event-based visual odometry/simultaneous localization and mapping.
\end{IEEEbiography}

\begin{IEEEbiography}[{\includegraphics[width=1in,height=1.25in,clip,keepaspectratio]{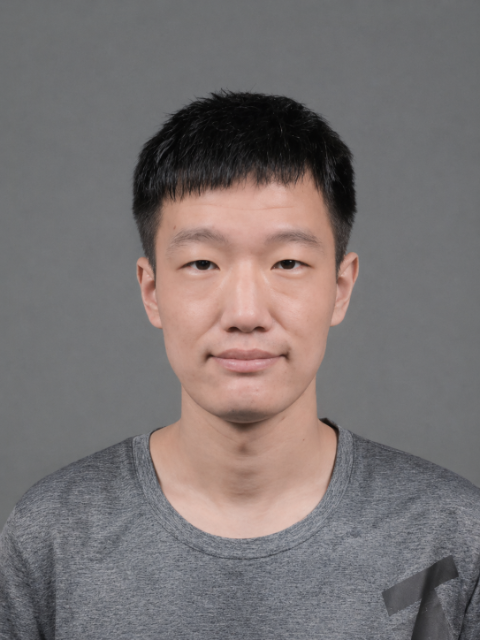}}]{Delei Kong}
(Graduate Student Member, IEEE) received the B.S. degree in automation from Henan Polytechnic University, Zhengzhou, China, in 2018, and the M.S. degree in control engineering from Northeastern University, Shenyang, China, in 2021.
He was an Algorithm Intern with SynSense Tech. Company Ltd., Chengdu, China (2020-2021), and an R\&D Engineer (Advanced Vision) with the Machine Intelligence Laboratory, China Nanhu Academy of Electronics and Information Technology (CNAEIT), Jiaxing, China (2021-2022).
Since 2022, he has been pursuing the Ph.D. degree in control science and engineering at Hunan University, Changsha, China.
His research interests include event-based perception and state estimation, event-aided image restoration, and neuromorphic robotics.
\end{IEEEbiography}

\begin{IEEEbiography}[{\includegraphics[width=1in,height=1.25in,clip,keepaspectratio]{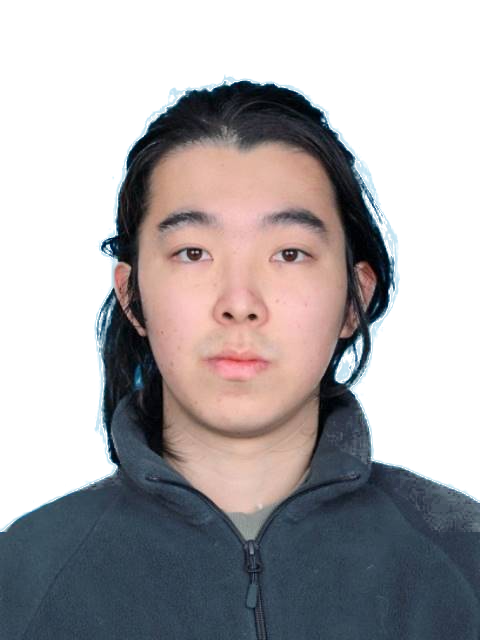}}]{Xin Wang}
received the B.S. degree in Software Engineering from the University of Electronic Science and Technology of China, Chengdu, China in 2025.
He is currently working toward the Ph.D. degree in the Neuromorphic Automation and Intelligence Lab (NAIL) with School of Robotics at Hunan University.
His research interests primarily focus on event-based visual odometry/simultaneous localization and mapping.
\end{IEEEbiography}

\begin{IEEEbiography}[{\includegraphics[width=1in,height=1.25in,clip,keepaspectratio]{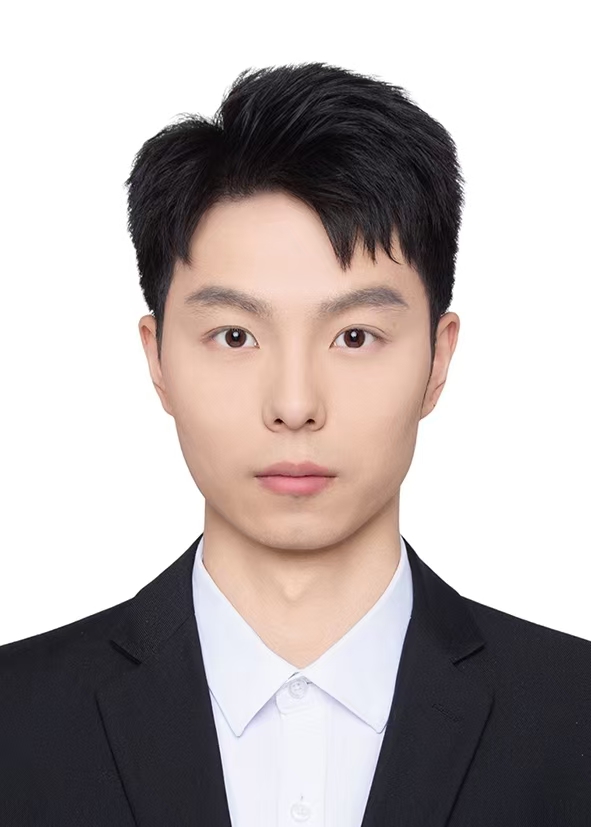}}]{Kaizhen Sun}
received the B.S. degree in Automation from Nanjing University of Aeronautics and Astronautics (NUAA), Nanjing, China, in 2023, and the M.S. degree in Control Science and Engineering from Hunan University, Changsha, China. He is currently pursuing the Ph.D. degree at the School of Intelligence Science and Technology, Nanjing University, Nanjing, China. His research interests include event-aided time-to-collision (TTC) estimation and embodied intelligence.
\end{IEEEbiography}

\begin{IEEEbiography}[{\includegraphics[width=1in,height=1.25in,clip,keepaspectratio]{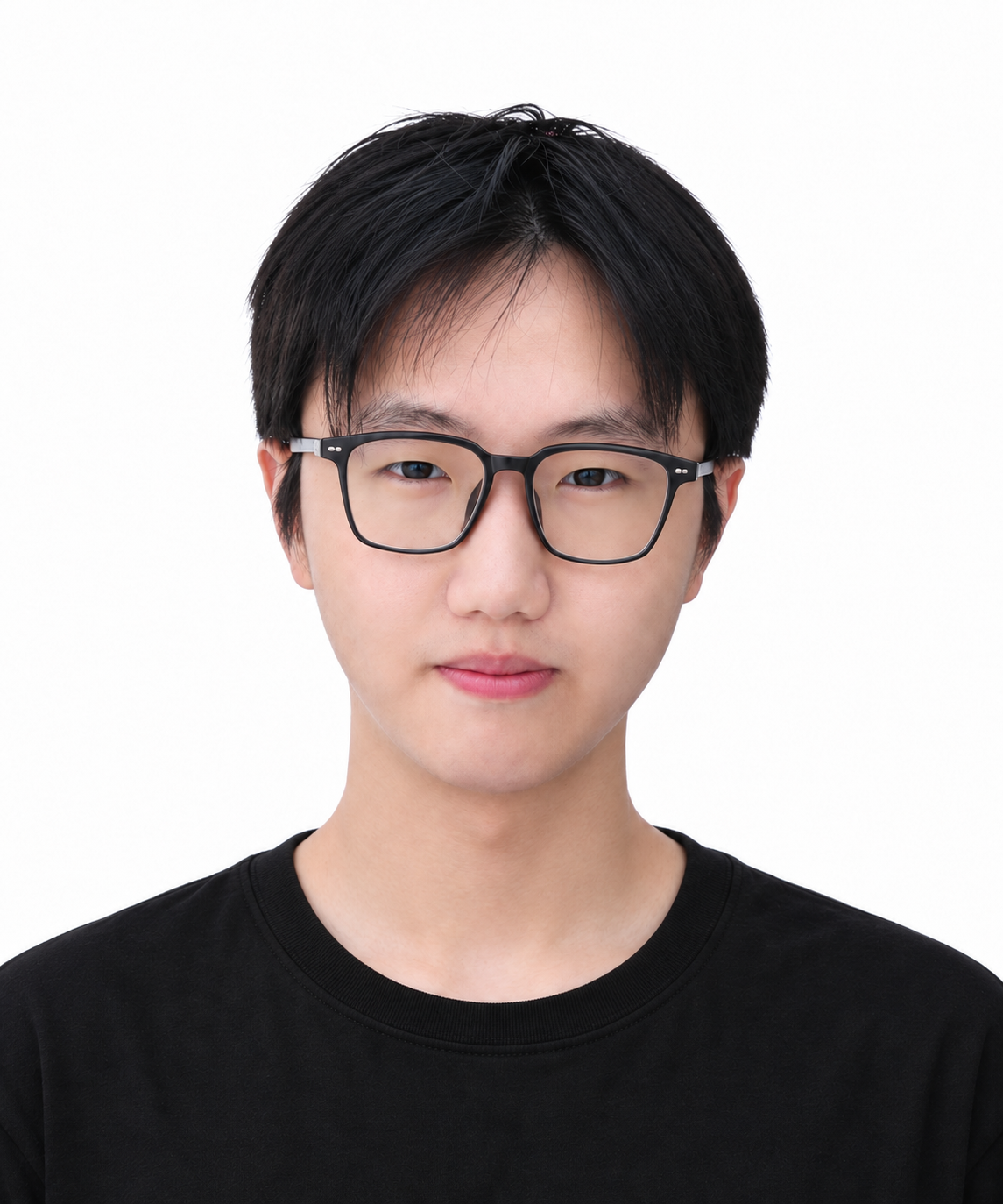}}]{Yusen Zhang}
is currently working toward the B.Eng. degree in Automation at the College of Artificial Intelligence and Robotics, Hunan University, Changsha, China. His research interests include computer vision, visual object tracking, and LED identification.
\end{IEEEbiography}

\begin{IEEEbiography}[{\includegraphics[width=1in,height=1.25in,clip,keepaspectratio]{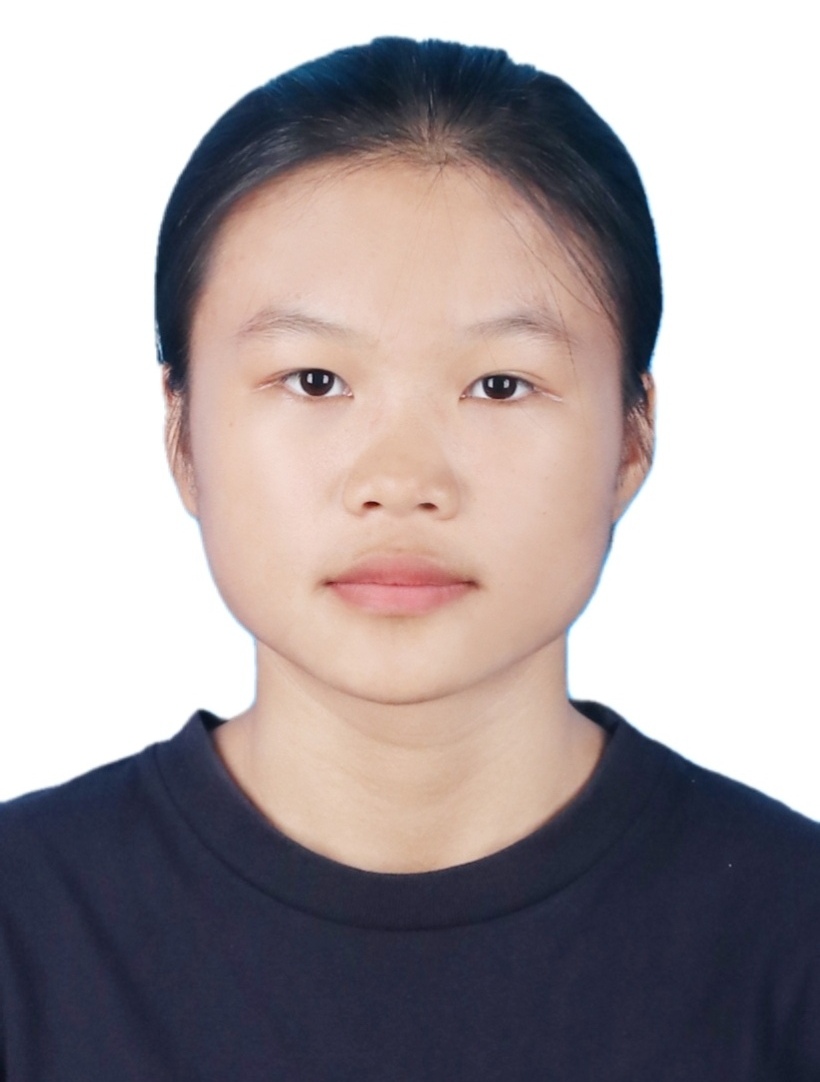}}]{Mingzhen Gan}
is currently working toward the B.Eng. degree in Automation at the College of Artificial Intelligence and Robotics, Hunan University, Changsha, China.
Her research interests include embodied navigation and robot learning.
\end{IEEEbiography}

\begin{IEEEbiography}[{\includegraphics[width=1in,height=1.25in,clip,keepaspectratio]{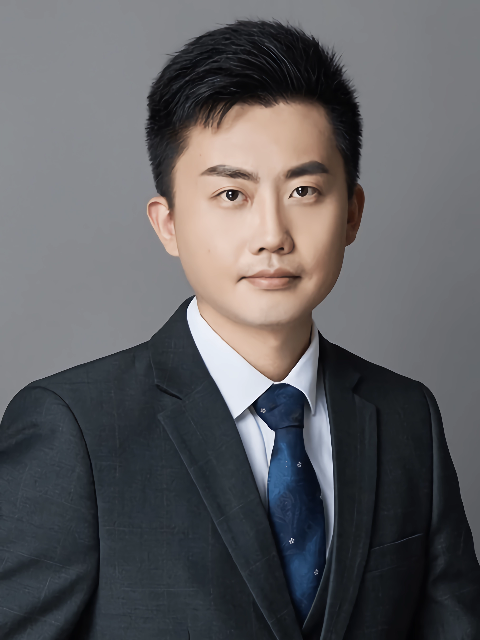}}]{Yi Zhou}
(Member, IEEE) is a Professor at Hunan University, where he directed the Neuromorphic Automation and Intelligence Lab (NAIL).
He obtained his Ph.D. degree in engineering and computer science from the Australian National University, Canberra, Australia in 2018.
He was a visiting scholar at ETH Zurich (2017-2018) and was awarded the NCCR Fellowship Award by the Swiss National Science Foundation for his research on neuromorphic event-based 3D vision. 
From 2019 to 2021, he was a postdoc research fellow at the HKUST-DJI Innovation Joint Lab, where he proposed the world's first open-source event-based stereo visual odometry (ESVO) system.
His research interests include spatial intelligence, geometric computer vision, and dynamic vision sensors. 
His team has received several academic honors, including first place in the SLAM Challenge at the CVPR 2025 Workshop on Event-Based Vision in Nashville, selection as a CVPR 2025 Paper Award Candidate, and recognition as a WRC 2026 SARA Best Paper Award Finalist, etc.
He is currently serving as an associate editor for IEEE RA-L and IEEE T-RO.
\end{IEEEbiography}

\end{document}